\documentclass[10pt,twocolumn,letterpaper]{article}

\usepackage{wacv}              

\usepackage{graphicx}
\usepackage{amsmath}
\usepackage{amssymb}
\usepackage{booktabs}
\usepackage{dirtytalk}
\usepackage{multirow}
\usepackage{xspace}
\usepackage{hyperref}

\usepackage[capitalize]{cleveref}
\crefname{section}{Sec.}{Secs.}
\Crefname{section}{Section}{Sections}
\Crefname{table}{Table}{Tables}
\crefname{table}{Tab.}{Tabs.}

\newcommand{\sysName}{PGRID}
\newcommand{\repoName}{\href{https://github.com/USAFORUNHCRhive/turkana-grid-mapping}{turkana-grid-mapping}}
\newcommand{\partnerName}{USA for UNHCR and HOTOSM}

\begin{document}
\title{PGRID: Power Grid Reconstruction in Informal Developments Using High-Resolution Aerial Imagery}
\author{
Simone Fobi Nsutezo\textsuperscript{1}\thanks{Corresponding author: \texttt{sfobinsutezo@microsoft.com}}, \enskip Amrita Gupta\textsuperscript{1}, \enskip Duncan Kebut \textsuperscript{2}, \enskip Seema Iyer \textsuperscript{3},\\ Luana Marotti\textsuperscript{1}, \enskip Rahul Dodhia\textsuperscript{1}, Juan M. Lavista Ferres\textsuperscript{1},\enskip Anthony Ortiz\textsuperscript{1} \\ \\
    \textsuperscript{1}Microsoft AI for Good Research Lab, \enskip \textsuperscript{2} HOTOSM, \\
    \textsuperscript{3}USA  for UNHCR
}
\maketitle

\begin{abstract}
 As of 2023, a record 117 million people have been displaced worldwide, more than double the number from a decade ago \cite{unhcr_displaced}. Of these, 32 million are refugees under the UNHCR's mandate, with 8.7 million residing in refugee camps. A critical issue faced by these populations is the lack of access to electricity, with 80\% of the 8.7 million refugees and displaced persons in camps globally relying on traditional biomass for cooking and lacking reliable power for essential tasks such as cooking and charging phones. Often, the burden of collecting firewood falls on women and children, who frequently travel up to 20 kilometers into dangerous areas, increasing their vulnerability.\cite{bonyan}
 
Electricity access could significantly alleviate these challenges, but a major obstacle is the lack of accurate power grid infrastructure maps, particularly in resource-constrained environments like refugee camps, needed for energy access planning. Existing power grid maps are often outdated, incomplete, or dependent on costly, complex technologies, limiting their practicality. To address this issue, {\sysName} is a novel application-based approach, which utilizes high-resolution aerial imagery to detect electrical poles and segment electrical lines, creating precise power grid maps. {\sysName} was tested in the Turkana region of Kenya, specifically the Kakuma and Kalobeyei Camps, covering 84 km² and housing over 200,000 residents. 
 
Our findings show that {\sysName} delivers high-fidelity power grid maps especially in unplanned settlements, with F1-scores of 0.71 and 0.82 for pole detection and line segmentation, respectively. This study highlights a practical application for leveraging open data and limited labels to improve power grid mapping in unplanned settlements, where the growing number of displaced persons urgently need sustainable energy infrastructure solutions.
\end{abstract}

\section{Introduction}
\label{sec:intro}
Reliable electricity access in refugee camps is critical to supporting everyday tasks such as cooking, phone charging and powering small businesses. Electricity consumption and income/earning potential are tightly correlated over time\cite{mem}, where increased electricity consumption is in lockstep with increased income. Thus electricity access in camps may better support residents' ability to earn a living.

Many refugee camps are often found in remote areas with little to no access to electricity. As a result, about 8.7 million people in refugee camps do not have access to much needed electricity\cite{bonyan}, to improve their livelihoods. Thus mapping the intricate layout of power grids is important for quantifying electricity access within camps and for strategic planning to support further electrification of camps.

High resolution information on power grid layout/structure is scarce, partial and often outdated. Advanced sensing devices, such as Advanced Metering Infrastructure (AMIs) and Phasor Measurement Units (PMUs), offer a pathway to grid mapping \cite{luan2015smart}. These technologies are capable of delivering real-time data on voltage measurements, thereby supporting voltage comparisons across meters to recreate the distribution grid. However, deploying smart meters in refugee camps across the distribution network may come with financial and technical challenges. The cost of these devices across the network and the specialized knowledge required to interpret the data they provide, particularly in terms of power-flow physics to map power grid layouts, makes them infeasible for unplanned settlements.

\textbf{Mapping power grids from remotely sensed imagery.}
As an alternative to mapping power grids from smart meter devices, satellite and aerial imagery has been utilized. Early work by Development Seed \cite{developmentseedhv} demonstrated a human-in-the-loop approach to mapping high voltage (HV) infrastructure. This approach used high resolution 50 cm/pixel optical satellite imagery to first detect HV towers, and then a team of experienced mappers  traced HV lines between towers. Alternatives to this workflow leverage lower resolution Synthetic Aperture Radar (SAR-C) imagery to detect HV steel pylons \cite{developmentseedsar}. This work indicated that grid infrastructure can be identified in satellite imagery, given the right image resolution and power infrastructure material visible in the image.  While HV lines provide information about the transmission network, they do not provide a view of the distribution network which is more reflective of where electricity is consumed at the residential and commercial levels. Moreover, the distribution network captures the various electric connection points and the ease and cost of making a power connection \cite{fobi2021scalable}. \cite{mvmapping} show that identifying the distribution network from satellite imagery remains challenging as distribution networks typically consist of smaller poles and lines, which can be hard to detect even at 50cm/pixel. As an alternative, Visible Infrared Imaging Radiometer Suite (VIIRS) nighttime lights is used in combination with network algorithms (e.g. Dijkstra's shortest path) to estimate the distribution network given electrified settlements. While this approach approximates the grid, it does not provide a true representation of the network and is sensitive to noise in VIIRS data such as flares, background lighting and temporal changes in luminosity\cite{arderne2020predictive}. GridTracer is an approach to mapping power grids from overhead imagery using deep learning \cite{huang2021gridtracer}. Alongside GridTracer, a benchmark dataset of overhead imagery, towers and lines was released to evaluate power grid mapping algorithms. Our work is most similar to GridTracer in its usage of overhead imagery, thus it serves as a baseline for comparison. Streetview imagery has also been used as an alternative to overhead imagery to map grid layouts \cite{tang2019fine}. While this provides an interesting application of streetview imagery, this approach is challenging to implement in unplanned settlements as streetview imagery is often not available for these locations. 

\begin{figure}[t]
\includegraphics[scale=0.45]{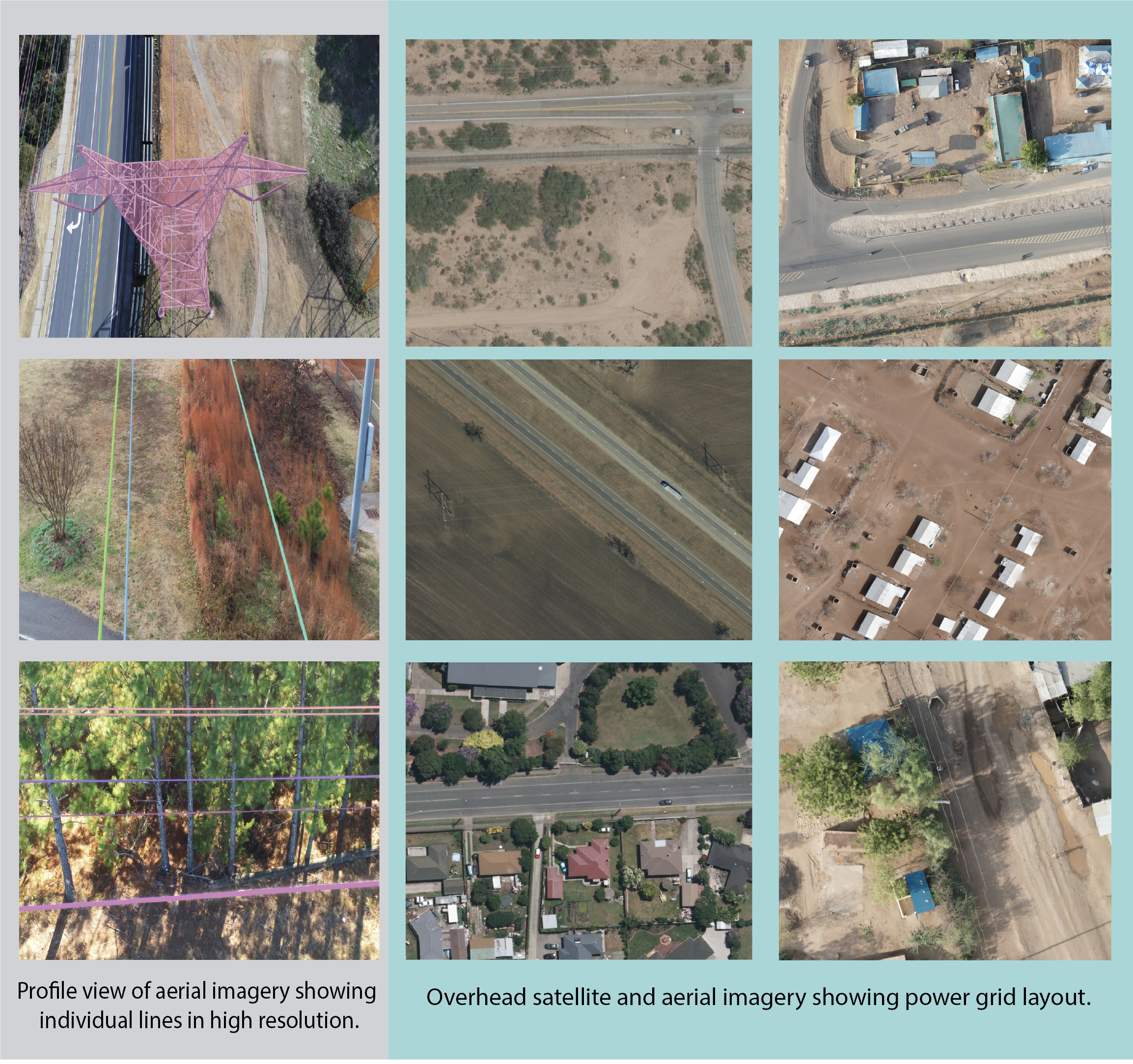}
\caption{Profile views (left) better support mapping individual lines and tower structures, while overhead imagery (right) better supports mapping the layout of a power grid.}
\label{fig:drone_comparisons}
\end{figure}

More recently, higher spatial resolution data collection through Unmanned Aerial Vehicles (UAV) such as drones has become more accessible with open source platforms such as OpenAerialImagery. As a result, drone imagery is increasingly utilized to map grid infrastructure. Using the Transmission Towers and Power Lines (TTPLA) dataset\cite{abdelfattah2020ttpla}, \cite{an2023duformer} propose a transformer based segmentation model to map individual transmission lines and towers. While these works advance the field of grid mapping from drone imagery, the data used in algorithm development consists of high voltage transmission lines at very high  resolution, with profile views as opposed to overhead spatial views (See Figure \ref{fig:drone_comparisons}). Datasets in profile view, are better suited for mapping individual lines and tower structures as opposed to reconstructing the entire grid structure from the dataset.

\textbf{Application context.}
According to the UN High Commission on Refugees (UNHCR), by mid-2023 there were approximately 32 million refugees under UNHCR mandate\cite{unhcr_displaced}, with about 22\% (likely a lower bound) of the refugee population living in refugee camps\cite{refugeecamp_nums}. This resulted in the creation of several informal settlements to provide safe shelter and amenities for displaced populations. An important aspect to supporting resettlement and enhancing quality of life for communities within these settlements is access to reliable and sufficient power. Accurate and updated maps of the power distribution infrastructure can inform relief organizations and field operators about power infrastructure needs and better support planning, resource allocation and operations within the camp. Given the nature of informal settlements, power distribution grids more likely appear overhead and are less organized, as the power grid is established to align with housing locations, which may be more unplanned compared to planned settlements. In this paper, in collaboration with \partnerName, we deploy our approach to mapping overhead power distribution infrastructure in the Kakuma Camp and Kalobeyei Integrated Settlement covering 84 km$^2$, and home to over 200,000 residents\cite{turkana_pop}.    
{\sysName} is evaluated using manually tagged labels from well-trained mappers. We also create a geo-referenced power grid layout map and show that our approach can be used to augment existing power grid maps.

\textbf{Contributions.}
This work presents an approach to mapping power grid distribution infrastructure from overhead aerial imagery at a spatial resolution of $\sim$6cm/pixel. The method is applied in practice to mapping refugee camps thereby enabling humanitarian mappers to rapidly map grid infrastructure in ever growing camps. In this work, we propose pole detection and line segmentation approaches for mapping low and medium voltage distribution lines. The main contributions of this paper are: 
\begin{enumerate}
    \item Easy-to-deploy algorithms for detecting electrical poles using point supervision and segmenting electrical lines. Our pole detection method simplifies the label acquisition process, as point labels are quicker and easier to obtain compared to bounding boxes.
    \item Deployment of our {\sysName} system in an unplanned settlement where we show high model performance in reconstructing the layout of the existing power grid .
    \item We demonstrate that our power distribution grid map enhances existing power grid estimates, and can serve as a viable approach to obtaining high resolution maps.
\end{enumerate}
We share our github repository and demo here: {\repoName}.
\section{Problem Formulation}
\label{sec:formulation}
For the task of mapping power distribution grids from overhead high resolution drone imagery, we formulate the problem as follows: for a given area $a$, the unknown power distribution grid network ($\mathcal{G}_a$) is defined as an undirected acyclic graph $\mathcal{G}_a(p,l)$, where $p$ is a set of poles ($p = {p_1,p_2, p_3,...p_n}$) and $l$ are the line segments connecting the poles ($l = {l_{12},l_{13},l_{34},...l_{mn}} = {(p_1,p_2),(p_1,p_3),(p_3,p_4),...(p_m,p_n)} $). Under perfect conditions, a power distribution grid can be fully mapped using accurate information about all line segments in the network. However, we observe from drone imagery that electrical lines appear in the image as very thin lines and their visibility within the image is dependent on the time when the imagery is taken, the lens angle and the building density within the image. Moreover, some areas may still be undergoing construction with existing poles but no lines. Because electrical lines may not always be  visible, we leverage complementary information about electrical poles and detect both electrical lines and poles.
Thus the objective of this paper is to estimate a power distribution grid ($\hat{\mathcal{G}}_a$) in a new area using geospatial machine learning to predict the location of electrical poles and lines that constitute the power distribution grid, given points and lines from a known area. 
\begin{equation}
    \hat{G}_a = M(\mathcal{P}(X), \mathcal{L}(X))
\end{equation}
$\hat{\mathcal{G}}_a$ is obtained in a two-step process as follows: First, we train a pole detection model ($\mathcal{P}_a$), given input imagery ($\mathcal{X}_a$) to accurately detect electrical poles given point pole locations from a known area. Next, we train a line segmentation model ($\mathcal{L}_a$) also given input imagery ($\mathcal{X}_a$) to detect connecting lines. Detected poles and lines are merged after post-processing to recreate the power distribution grid $\hat{\mathcal{G}}_a$.

\section{{\sysName}: Reconstruction of grid layouts.}
\label{sec:methods}
The {\sysName} method entails three steps: i) a pole detection model to detect electrical pole locations within the image ii) a line segmentation model to segment electrical lines from imagery and iii) a post-processing step to merge detected poles and line segments to obtain a unified layout. We outline each step of our methodology below.

\subsection{Electrical Pole Detection}
Accurate identification of electrical poles is performed using our proposed pole detection model. We deploy a Fully Convolutional Network (FCN8) \cite{long2015fully} semantic segmentation model to detect poles within the image. We build on the network architecture and multi-component loss function proposed by \cite{laradji2018blobs} to count and localize cars and adapt it to detect electrical poles. A labeled dataset of point locations is used to indicate electrical pole locations. This labelled dataset, coupled with corresponding images is used as a training signal to learn pole representations within the imagery (see supplementary material for illustration of pole detection network). The model is trained with a four component loss: 
i) \textit{image-level loss}, is a negative log-likelihood loss which encourages the model at the image-level to detect electrical poles. For images that have poles, the loss increases the likelihood that at least one pixel is predicted as a pole, while decreasing the likelihood of predicting pixels as poles when there are no poles in the image. Here all image pixels are considered, ii) \textit{point-level loss}, which serves as a localization loss, helping the model to accurately learn pole structures within the image. A negative log-likelihood loss is applied only to the annotated pole pixels to encourage the model to correctly identify poles. Both the image and point level losses evaluate the likelihood that a pixel is in the pole class. The next two losses, evaluate the likelihood that a pixel belongs to the background class. The iii) \textit{split-level loss}, encourages the model to predict unique blobs for all annotated poles. First, boundaries between ground truth poles are found using a watershed segmentation algorithm\cite{beucher2018morphological}. These boundaries are annotated as background while the area within the boundaries are annotated as foreground. The model learns to predict the probability that a pixel belongs to the background class. Here we also use a negative log-likelihood loss,  weighted by the number of poles within the blob. This encourages the model to predict boundaries such that only one pole is in each blob. iv) \textit{false positive loss}, which discourages the model from making false positive predictions. A negative log-likelihood loss is applied to pixels with no ground truth pole annotations, thereby minimizing the false positive errors by the model. 

\begin{figure}[t]
\centering
\includegraphics[scale=0.65]{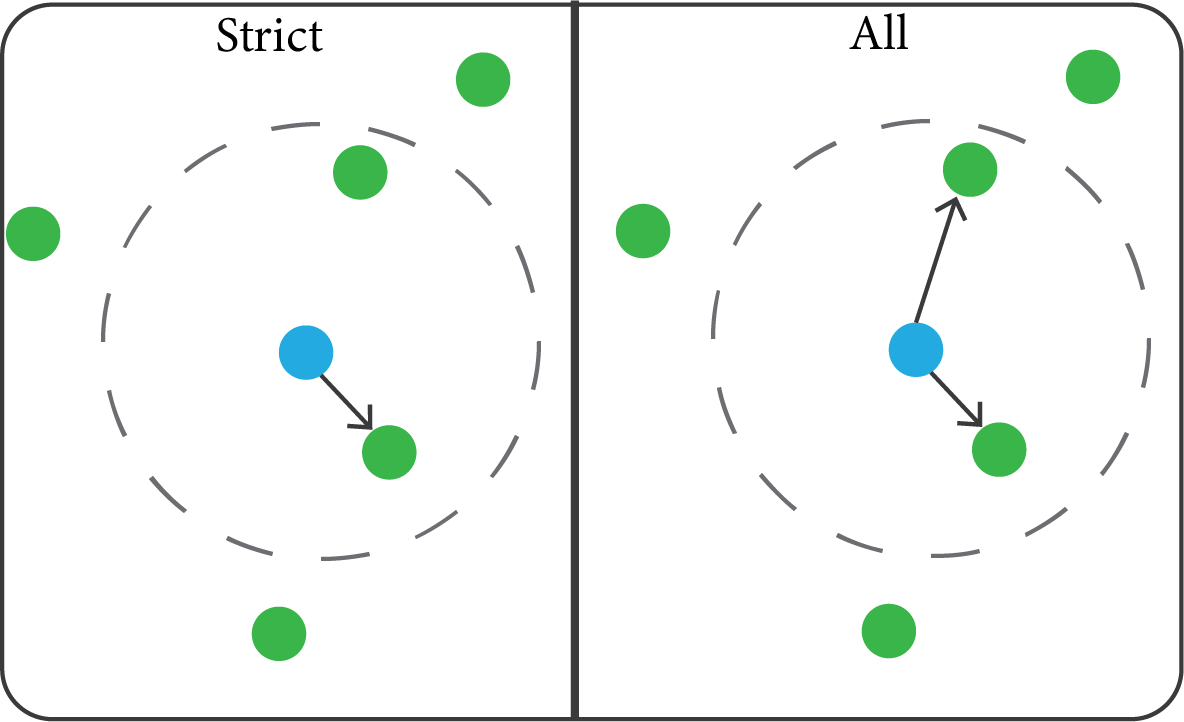}
\caption{Illustration of \textit{strict} (left) and \textit{all} (right) matching variants for evaluating performance of pole predictions. Ground truth poles are shown at the center in blue, with predicted poles in green surrounding the ground truth pole. The dashed circle represents the threshold ($th$). In the \textit{strict} match variant, the ground truth pole is matched to its closest prediction, so long as the prediction is within the threshold. In the \textit{all} variant, the ground truth pole is matched to all predicted poles within the threshold. The \textit{all} match is a better approach to evaluate performance for cases where mask predictions for poles is a non-continuous blob.}
\vspace{-4pt}
\label{fig:matching_visual}
\end{figure}
A small learning rate of 1e-6 was used to train the pole detection model. Image augmentation techniques such as vertical and horizontal flips, color jittering and random rotations were used as a form of regularization during training.

\textbf{Pole Detection Metrics.}
Electrical poles are represented as points in the label dataset, and predictions obtained from the model are vectorized and their centroids taken as predicted pole locations. Matching ground truths with predictions cannot be performed using exact matches given the high precision nature of points. Thus, we set an acceptable distance thresholds $th$ and utilize two matching variants known as \textit{strict} and \textit{all} matches, as illustrated in Figure \ref{fig:matching_visual}. In the \textit{strict} variant, we perform a one-to-one match between the ground truth pole and a predicted pole, given both poles are within the acceptable distance. When multiple predictions occur within the acceptable distance, we select the closest prediction to the ground truth pole. In the \textit{all} variant, we perform a one-to-many match between the ground truth and predicted poles, given $th$. The pole matching variants are used to compute the following metrics:

\textit{Distance-based Precision.} Distance-based precision represents the proportion of predicted poles that have a corresponding ground truth pole, given $th$. Distance-based precision is computed given the modified precision formula: 
\begin{equation}
    Precision_{th} =\frac{TP_{th}}{(TP_{th}+FP_{th})}
\end{equation}
where $TP_{th}$ represents true positives at $th$, and $FP_{th}$ represents false positives.
The \textit{strict} match represents a one-to-one match between predicted poles and ground truth poles, where the closest predicted pole is matched to a ground truth pole, given $th$. Additional predicted poles within the designated threshold are considered false positives. In the \textit{all} match, all predicted poles within the designated threshold $th$ are matched to the ground truth pole. 

\textit{Distance-based Recall.} Distance-based recall represents the proportion of poles that are correctly predicted as poles, given $th$. Thus the modified recall formula is:
\begin{equation}
    Recall_{th} =\frac{TP_{th}}{(TP_{th}+FN_{th})}
\end{equation}

where $TP_{th}$ represents true positives at $th$, and $FN_{th}$ represents false negatives. $TP_{th}$ is defined as the number of ground truth poles that are correctly predicted as poles, given a distance threshold ($th$).

\textit{Distance-based F1-Score.} Distance-based F1-score represents the harmonic mean between precision and recall, given specific matching distance threshold ($th$).
\subsection{Electrical Line Segmentation}
We propose patch-level segmentation using an asymmetric DeepLabV3 \cite{chen2017rethinking} architecture (see supplementary materials) to segment electrical lines. We apply a scaling factor (\textit{sf}) to the ground truth label raster to obtain patch-wise labels, indicating line presence or absence. This approach is selected over pixel-level segmentation because of the very thin nature of electrical lines. As an alternative, the scaling factor allows the model to classify each patch as having an electrical line or not. \textit{sf} of 1 is equivalent to pixel-level segmentation, while \textit{sf} with a value the same as the image size approximates image classification. Through experimentation, we observe that an \textit{sf} of 4 yields good detection and localization of line segments. At this \textit{sf} the ground truth mask is reduced to 1/4 of its original size e.g. 512 x 512 groundtruth mask becomes a 128 x 128 groundtruth after scaling, and the model classifies every 4x4 image patch as containing a line or not. DeepLabV3 is selected due to its atrous convolutions which support rich contextual feature extraction, a desirable attribute for line segmentation. 

The line segmentation model is trained as a binary model, with a cross-entropy loss and an Adam optimizer at a learning rate of 1e-5. Other loss functions, such as Dice and Focal loss, yielded comparable performance to cross-entropy so no further optimization was done. Similar to the pole detection model, the line segmentation model is trained with augmented data samples to prevent overfitting. We evaluate the line segmentation model at the pixel level (buffered by 2 meters on each side) and compare it to the ground truth labels. Buffering is done because the ground truth test labels weakly align with visible electrical lines in the imagery as opposed to training labels that are highly aligned with visible electrical lines in the image. As a result, the model is trained to predict visible electrical lines in the imagery rather than infer connections between electrical poles. We observe that ground truth labels in the test set have an offset of $\sim$2m from visible electrical lines within the imagery. In addition to reporting the mean intersection-over-union (mIOU), precision, recall and F1-score are also reported. For grid layout reconstruction, identifying grid presence/absence is more informative than having high fidelity alignment between the predictions and ground truth, especially given weak ground truth line labels. Thus we measure model accuracy in addition to alignment. 

\begin{figure}[t]
\centering
\includegraphics[scale=0.5]{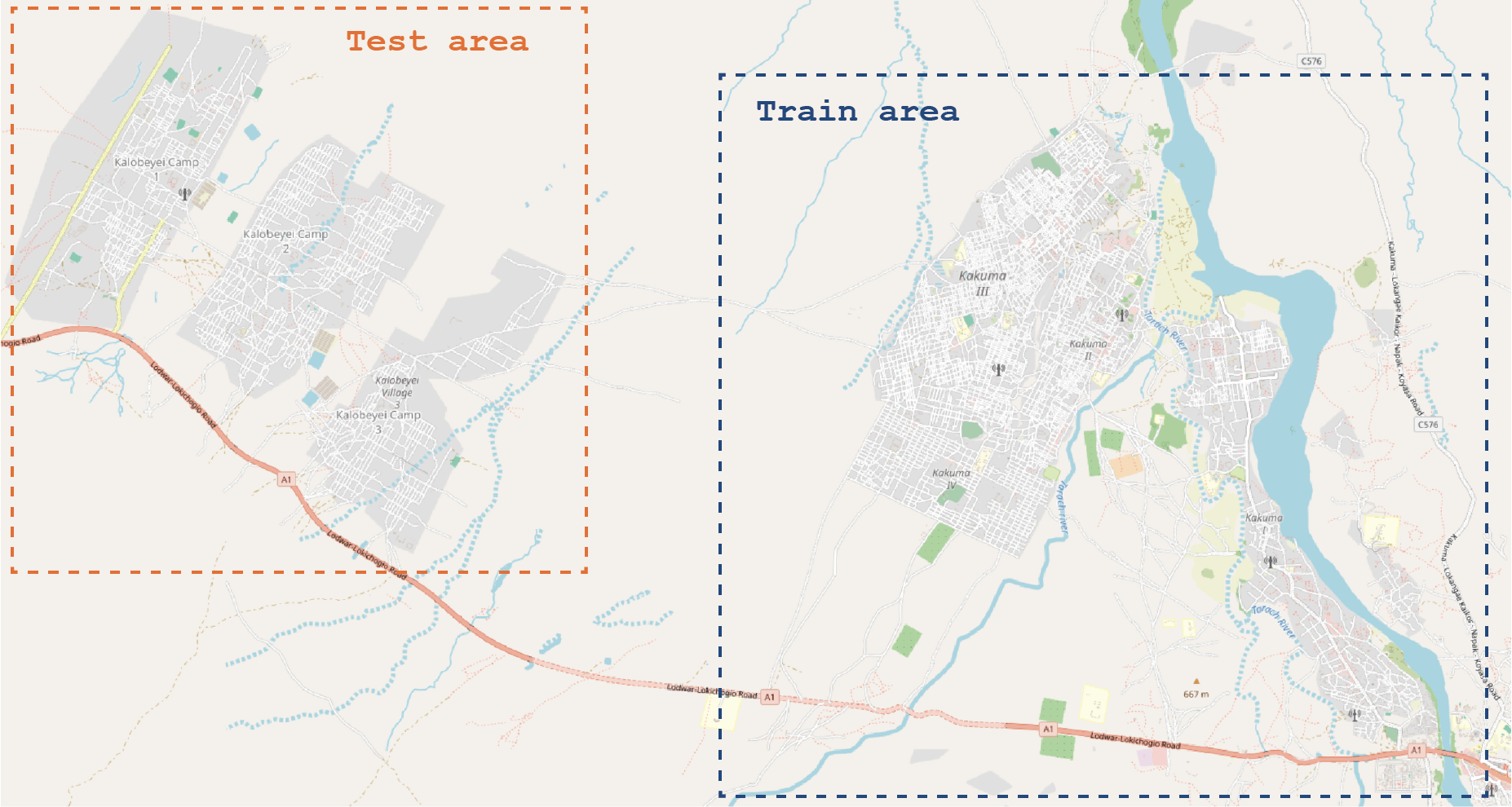}
\caption{Spatially distinct geographic train and test splits for model training. Models were trained on the Kakuma Camp (right) and tested on the Kalobeyei Camp (left).}
\label{fig:geosplits}
\end{figure}
\subsection{Hard Negative Mining}
Learning to detect electrical poles from imagery can be a difficult task as objects such as fences, street lights, billboard poles and even trees, would be closer in representation to electrical poles, compared to the background. Thus, these objects will be falsely classified as electrical poles thereby making reconstruction of the grid layout difficult. To address this issue, we implement a hard negative mining strategy for pole detection, where difficult negative samples such as fences are labelled and included in the dataset for training. By showing the model hard negative examples, the model learns distinct representations for electrical poles thereby further reducing false positive predictions.

\subsection{Unified Power Grid Layout}
After training both the pole detection and line segmentation models, the predicted pole raster files are vectorized to obtain polygons, and the polygon centroids are taken as pole locations. For the final grid map, we filter out small polygons prior to generating centroids. Predicted power lines are also vectorized, skeletonized and the same 2 m buffer is applied around the line skeletons. The buffered line predictions are then polygonized to obtain electrical line predictions. This provides a unified power grid ($\hat{\mathcal{G}}_a$) in an easily accessible and lightweight vectorized format for humanitarian mappers to use as visual guides when mapping grids.

\section{Datasets}
\label{sec:data}
Mapping power grid infrastructure from imagery is contingent on having high resolution imagery with visible power infrastructure and corresponding ground truth annotations for model training. We utilize multiple datasets across diverse geographies to demonstrate the value of the contributions made in this paper.

\textbf{Benchmark Dataset.}
We use a benchmark dataset of overhead satellite imagery and corresponding pole tower and line annotations \cite{huang2021gridtracer} to evaluate the performance of our grid mapping algorithms. This dataset consists of overhead imagery collected from Arizona, USA (AZ); Kansas, USA (KS) and New Zealand (NZ). The images are provided at 0.3m, resampled from native resolutions ranging from 0.15m-0.3m. The first 20\% of imagery from each city is used for testing, while the 80\% is used for training and validation, consistent with the dataset authors' format. 

\textbf{Turkana Integrated Settlement.}
In November 2022, the United Nations High Commissioner for Refugees (UNHCR), Humanitarian OpenStreetMap Team (HOT), and Kenya Red Cross Society (KRCS) partnered to conduct drone mapping of Kakuma-Kalobeyei. Drones successfully covered 84 km$^2$, capturing imagery at $\sim$6 cm/pixel. The collected drone imagery was inspected by HOT mappers, who manually labeled electrical poles and lines for $\sim$20\% of the area. This paired dataset of manually annotated electrical poles and lines with drone imagery is used. Annotations for the Turkana Integrated Settlement camp were downloaded from OpenStreetMaps (OSM). The \textit{power} and \textit{line} tags were used to search OSM, where the following features were selected and downloaded: \textit{\say{pole}}, \textit{\say{line}} and \textit{\say{minor\textunderscore line}}. 

\begin{table}[b]
\caption{Distance based mean Average Precision (DmAP) for the tower detection task on the benchmark dataset. Comparison of GridTracer method with ours ({\sysName}). On average, {\sysName} tower detection outperforms GridTracer on the benchmark dataset.}
\begin{tabular}{l|lllll}
\hline
\textbf{Methods}    & \textbf{Backbone} & \textbf{AZ}   & \textbf{KS}   & \textbf{NZ}   & \textbf{Mean} \\ \hline
GridTracer & ResNet50 & 0.45 & 0.48 & 0.54 & 0.49  \\
{\sysName}  & ResNet50 & 0.57     & 0.75  & 0.55 & \textbf{0.62}  \\ \hline
GridTracer & ResNet101 & 0.73 & 0.53 & 0.59 & 0.62   \\
{\sysName} & ResNet101&    0.70  &   0.72   &  0.55    &   \textbf{0.66} 
\end{tabular}
\label{tab:pole_baseline}
\end{table}
To evaluate the generalizability of {\sysName}, the Turkana camps were geographically split as shown in Figure \ref{fig:geosplits}, where the model was trained with data from the Kakuma Camp (right) and tested in the Kalobeyei Camp (left). For the training data region, pole annotations from OSM were manually realigned to corresponding imagery prior to training the model. The line annotations from OSM were used as a guide to annotate actual lines in imagery as the lines obtained from OSM showed the general grid trajectory and not the actual electrical lines within the image. The models are trained and validated given the Kakuma train split and performance metrics are reported for the unseen Kalobeyei test split. 863 poles were used to train the pole detection model while 636 poles were used for evaluation. 171 line segments with a mean line length of 161 m were used to train the line segmentation model while 118 line segments with a mean line length of 240 m were used for evaluation.

\nopagebreak
\section{Experiments \& Results}
\label{sec:experiments}
\begin{table}[t]
\caption{Mean Intersection-Over-Union (mIOU) for electrical line segmentation on the benchmark dataset.}
\centering
\vspace{-1pt}
\begin{tabular}{l|llll}
\hline
\textbf{Methods}     &\textbf{AZ}   & \textbf{KS }  & \textbf{NZ}   & \textbf{Mean }\\ \hline
GridTracer-UNet& 0.50 & 0.34 & 0.38 &  0.41   \\
GridTracer-StackNetMTL& 0.54 & 0.40 & 0.47 & 0.46    \\
{\sysName}  & 0.51& 0.34 & 0.37 & 0.41  \\
\end{tabular}
\label{tab:line_baseline}
\end{table}

\subsection{Performance on benchmark dataset}
We evaluate the performance of our {\sysName} models on the benchmark dataset. 
Tower detection and line segmentation models are evaluated on the test data from all three regions. The data split is consistent with GridTracer's data split for evaluation. Table \ref{tab:pole_baseline} shows Distance-based mean Average Precision (DmAP) on the tower detection task for the benchmark dataset. We compare results from {\sysName} to the GridTracer algorithm. {\sysName} pole detection method using point labels outperforms the FasterRCNN object detection method utilized by the GridTracer algorithm. Using a ResNet50 architecture, {\sysName} outperforms the GridTracer tower detection method on all three locations. When the more expressive ResNet101 model is used, further performance gains are observed with {\sysName} outperforming by 19\% in Kansas and under-performing by only 3-4 \% in Arizona and New Zealand. In addition, {\sysName}  utilizes point labels, as opposed to bounding boxes, thereby making the label acquisition easier. \cite{bearman2016s} show that acquiring point annotations is faster and requires less effort than acquiring bounding boxes, thereby making point supervision more practical for real-world deployment. Line segmentation model performance on the benchmark dataset is presented in Table \ref{tab:line_baseline}. {\sysName} performs comparably to the GridTracer UNet model without the added complexity of predicting orientation with StackNetMTL\cite{batra2019improved}.

\subsection{Performance in Turkana Integrated Settlement}
{\sysName} pole detection and line segmentation models are run on the Turkana drone imagery and results are reported.
\begin{figure}[t]
\centering
\includegraphics[scale=0.55 ]{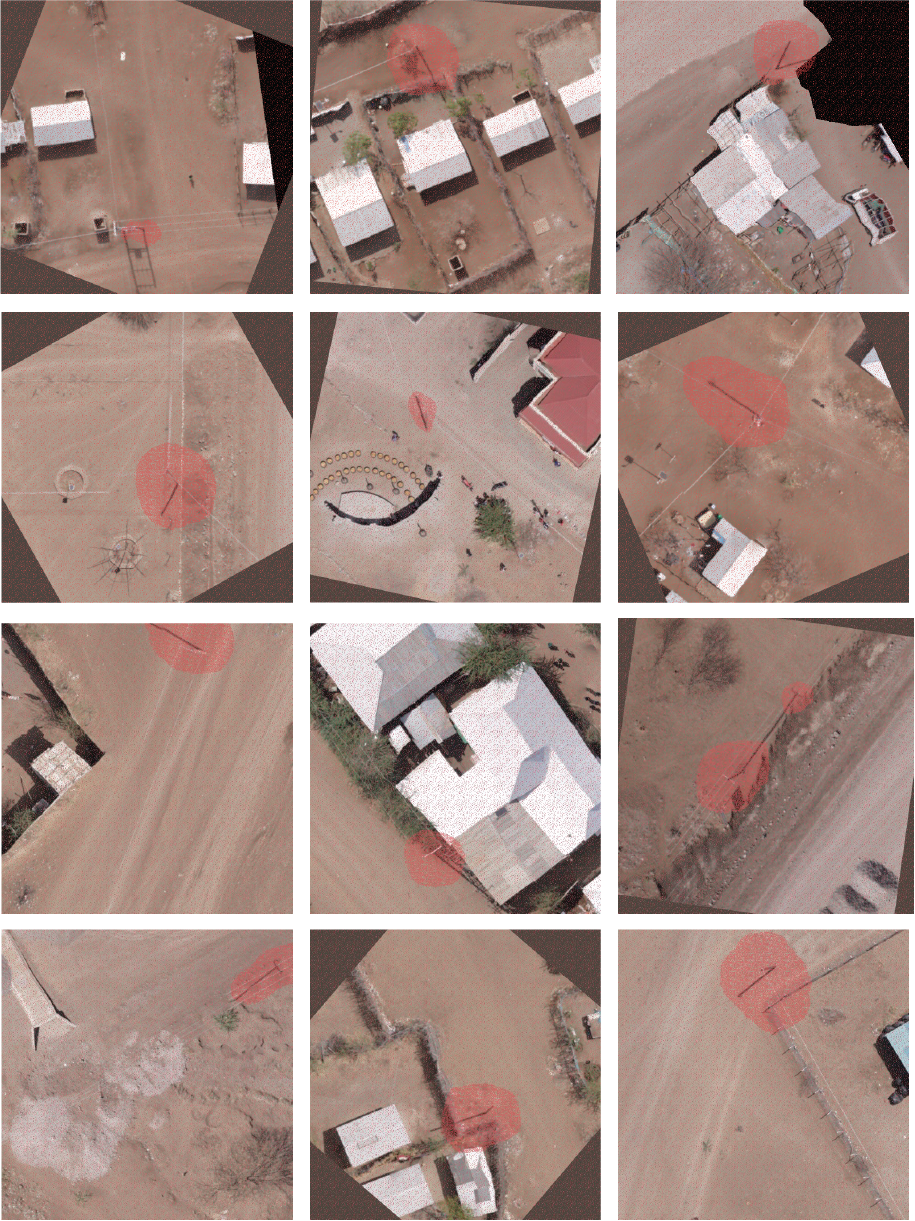}
\caption{Test set sample images with predicted poles from the trained pole detection model shown as red blobs.}
\label{fig:pole_results}
\end{figure}

\begin{table*}[t]
\caption{Pole detection results for 3 camps in the test region, given ResNet50 and ResNet101 architectures. Metrics are reported for a strict (one-to-one) match between the ground truth and predictions, and the all (many-to-one) match, given 3 distance thresholds.}
\resizebox{\textwidth}{!}{%
\begin{tabular}{l|lllll|lllll}
\hline
                      & \multicolumn{5}{c|}{\textbf{ResNet50  \&  10m}} & \multicolumn{5}{c}{\textbf{ResNet101 \& 10 m}}    \\ \hline
\multicolumn{1}{c|}{} & $P_S$     & $P_A$     & R      & $F1_S$   & $F1_A$   & $P_S$     & $P_A$     & R     & $F1_S$    & $F1_A$   \\ \hline
K1                    & $0.66\pm0.06$   & $0.70\pm0.05$   & $0.66\pm0.02$   & $0.66\pm0.03$  & $0.68\pm0.02$  &  $0.78\pm0.01$      &   $0.82\pm0.01$     & $0.63\pm0.01$      &     $0.70\pm0.01$   &  \textbf{\boldmath$0.71\pm0.01$}   \\
K2                    & $0.56\pm0.01$   & $0.62\pm0.02$   & $0.78\pm0.05$   & $0.65\pm0.02$  & $0.69\pm0.03$  &  $0.63\pm0.05$      & $0.68\pm0.03$       & $0.72\pm0.05$      &   $0.67\pm0.05$     &   \textbf{\boldmath$0.70\pm0.04$}    \\
K3                    & $0.63\pm0.03$   & $0.64\pm0.02$   & $0.80\pm0.04$   & $0.70\pm0.00$  & \textbf{\boldmath$0.71\pm0.00$}  &   $0.76\pm0.02$     &  $0.78\pm0.01$      & $0.62\pm0.06$      &    $0.68\pm0.04$    &   $0.69\pm0.04$    \\ \hline
\multicolumn{1}{c|}{} & \multicolumn{5}{c|}{\textbf{ResNet50    \& 7m}}   & \multicolumn{5}{c}{\textbf{ResNet101 \&  7 m}} \\ \hline
K1                    & $0.64\pm0.05$   & $0.65\pm0.05$   & $0.64\pm0.02$   & $0.64\pm0.02$  & $0.65\pm0.02$  &  $0.77\pm0.01$      &    $0.78\pm0.01$    &  $0.67\pm0.10$     &  $0.69\pm0.01$      &  \textbf{\boldmath$0.70\pm0.01$}     \\
K2                    & $0.54\pm0.02$   & $0.54\pm0.02$   & $0.76\pm0.04$   & $0.63\pm0.02$  & $0.63\pm0.02$  &   $0.63\pm0.05$     &  $0.64\pm0.04$      &   $0.71\pm0.05$    &    $0.66\pm0.05$    &  \textbf{\boldmath$0.67\pm0.04$}     \\
K3                    & $0.63\pm0.03$   & $0.63\pm0.03$   & $0.80\pm0.04$   & $0.70\pm0.00$  & \textbf{\boldmath$0.71\pm0.00$}  &   $0.75\pm0.01$     &  $0.77\pm0.01$      &   $0.62\pm0.06$    &  $0.68\pm0.04$      &  $0.69\pm0.04$     \\ \hline
\multicolumn{1}{c|}{} & \multicolumn{5}{c|}{\textbf{ResNet50 \&  5m}}  & \multicolumn{5}{c}{\textbf{ResNet101 \&  5m}}  \\ \hline
K1                    & $0.61\pm0.04$   & $0.62\pm0.04$   & $0.62\pm0.02$   & $0.62\pm0.02$  & $0.62\pm0.01$  &  $0.76\pm0.01$      & $0.76\pm0.01$       & $0.62\pm0.01$      &   $0.68\pm0.01$     &  \textbf{\boldmath$0.68\pm0.01$}     \\
K2                    & $0.52\pm0.03$   & $0.53\pm0.02$   & $0.74\pm0.06$   & $0.61\pm0.03$  & $0.61\pm0.03$  & $0.61\pm0.04$       & $0.61\pm0.04$       &  $0.69\pm0.05$     &   $0.65\pm0.05$     &  \textbf{\boldmath$0.65\pm0.05$}     \\
K3                    & $0.63\pm0.03$   & $0.63\pm0.03$   & $0.80\pm0.04$   & $0.70\pm0.01$  & \textbf{\boldmath$0.70\pm0.01$}  & $0.75\pm0.02$       &  $0.75\pm0.02$      &  $0.61\pm0.06$     &    $0.67\pm0.03$    &  $0.67\pm0.03$     \\ \hline
\end{tabular}%
}
\label{tab:turkana_poles}
\end{table*}

\textbf{Pole detection performance.}
Figure \ref{fig:pole_results} shows detected poles using the {\sysName} pole detection method for a small sample within the test set. Predicted poles are shown as red blobs while ground truth poles are visible within the images. This sample shows that point supervision results in detection of the full pole, sometimes with predicted blobs occurring on both edges of the pole (row 3, column 3). 
\begin{table}[b]
\caption{Comparison of F1-scores using ResNet101 architecture when model is trained with and without hard negative mining. We report results at a distance threshold of 10m.}
\begin{tabular}{l|lll|lll}
\hline
           & \multicolumn{3}{c|}{\textbf{$F1_S$}} & \multicolumn{3}{c}{\textbf{$F1_A$}} \\ \hline
 & K1      & K2     & K3     & K1       & K2      & K3      \\ \hline
No - HNM        & 0.70   & 0.64 & 0.60   & 0.72   & 0.66   &  0.61  \\
HNM       & 0.70   & 0.67 & 0.68  & 0.71  &  0.70   & 0.69    \\
\end{tabular}
\label{tab:hnm_results}
\end{table}

Table \ref{tab:turkana_poles} presents pole detection model performance for each of the three camps in the Kalobeyei test area, given ResNet50 and ResNet101 encoder backbones. Distance based Precision, Recall and F1-score for the \textit{strict} and \textit{all} match variants, across three trials are shown. Higher F1-scores are observed when using a ResNet101 backbone compared to a ResNet50 backbone. This aligns with expectation as the more expressive ResNet101 backbone learns better pole representations within the dataset. Predicted blobs many times encompass the whole pole, such that the centroid of the blob (used to compute precision) falls some distance away from the ground truth point (occurring at a pole edge). Thus, we evaluate performance at 3 distance thresholds (5m, 7m, 10m), as we observe that poles cast varying shadow lengths in the imagery (see supplementary material).
Improved detection is also observed as the distance thresholds are increased. When pole shadow lengths for a random sample of 100 poles in the test set are measured, less than 30\% of randomly sampled poles have a pole length under 5m. At this distance, viable detections are rejected for poles casting shadows larger than 5m. Thus at a distance threshold of 10m, even using the \textit{strict} one-to-one match, we observe improved performance for both ResNet backbones over a 5m and 7m. Note that 90\% of sampled poles cast shadows less than or equal to 10m. We also observe small variances between the \textit{strict} and \textit{all} match variants, indicating that the model is mostly making single detections for each ground truth pole. We considered bounding-box object detection methods such as FasterRCNN\cite{ren2016faster}, and observe that such object detection methods struggle to learn when a fixed sized bounding box is applied for the pole point labels. The uniformly sized label bounding boxes are either too small (leaving out a portion of the pole) or too big (including non-pole information) for a subset of the labels, thereby making it hard for the detection algorithm to consistently learn relevant pole features needed for meaningful detection. 
 
Detecting electrical poles from aerial imagery can be a non-trivial task, as multiple pole-like structures might be predicted as electrical poles, while these structures are in fact not electrical poles. Hard negative mining is applied to suppress false positives, where the model is trained with hard negative examples. Table \ref{tab:hnm_results} shows performance across the three camps with and without hard negative mining, at the 10m distance threshold using the \textit{strict} match and ResNet101 encoder. Hard negative mining improved model performance between 3-8 \% in K2 and K3.
\begin{figure*}[t]
\centering
\includegraphics[scale=0.6]{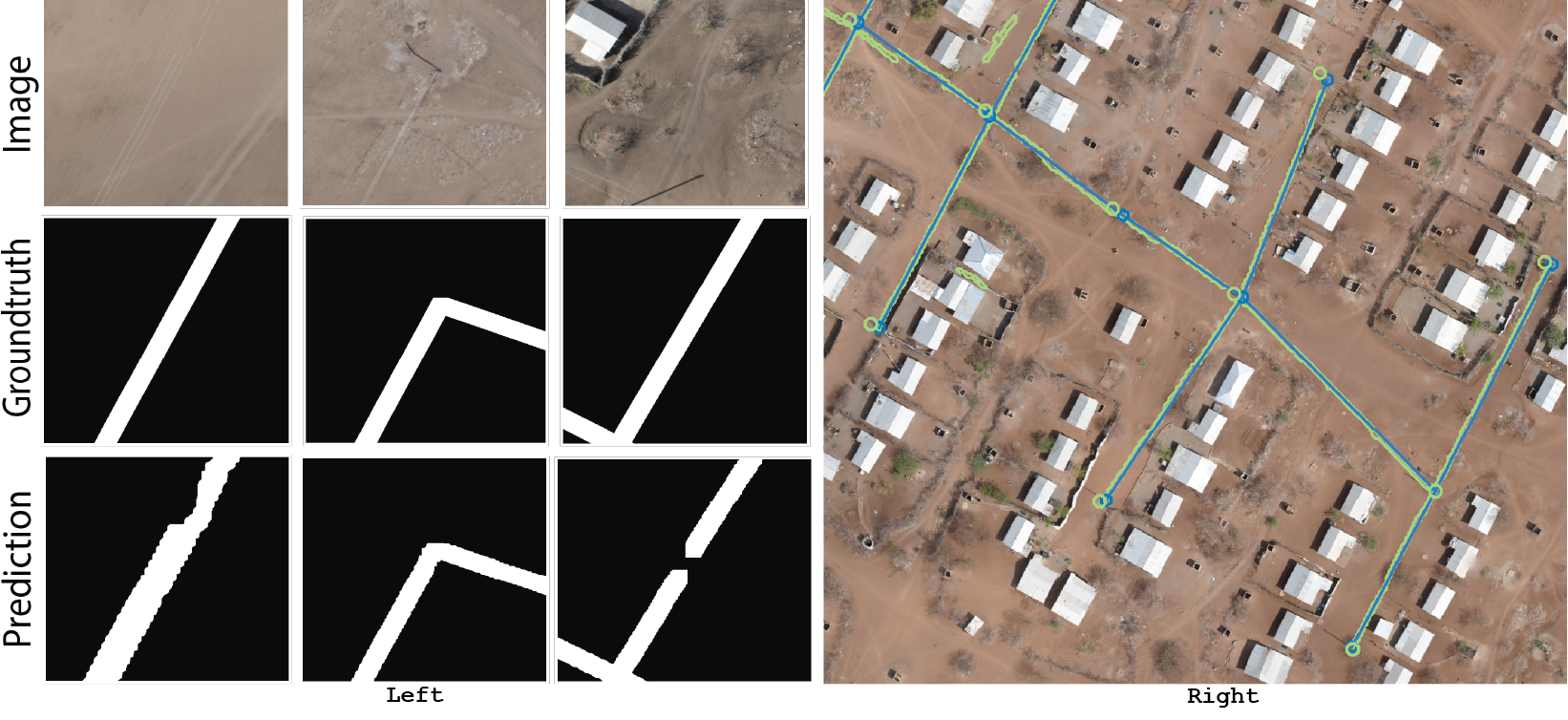}
\caption{\textbf{Left:} Sample images showing electrical lines (top), ground truth lines buffered by 2 m (middle) and model predictions (bottom). \textbf{Right:} Sample ground truth power line (blue) alongside on predicted power line (green). Electrical poles are shown as circles.}
\label{fig:line_results}
\end{figure*}

\textbf{Line segmentation performance.}
Table \ref{tab:line_ours} shows line segmentation model performance across the three camps in Kalobeyei. Mean Intersection-Over-Union (mIOU) is reported to measure alignment between the ground truth and predicted electrical lines, while precision, recall, F1-score are used to measure how well the model detects lines. For the purpose of mapping grid layouts, line detection is of higher importance when compared to perfect alignment.  

The {\sysName} line segmentation model is able to detect electrical lines with F1-scores between 0.77-0.82,  with the lowest performance occurring in K2. The lower performance in the K2 camp is because lines are drawn connecting poles while there are no visible lines in the imagery. Intuitively, connecting poles with lines makes sense however, we do not run a post-processing step connecting lines and poles, as our partners also observe that there are areas in the integrated settlement where the grid is still under-construction. Across 3 trials, there is small variance in model performance, indicating the stability of the model in segmenting electrical lines. 
\begin{table}[b]
\caption{Performance of line segmentation model, reporting alignment with the mIOU and detection accuracy with the F1-score.}
\resizebox{0.48\textwidth}{!}{%
\begin{tabular}{l|llll}
\hline
   & mIOU & Precision & Recall & F1-score \\ \hline
K1 & $0.70\pm0.01$ & $0.83\pm0.00$   &  $0.82\pm0.00$        &     $0.82\pm0.00$     \\
K2 &  $0.63\pm0.01$    &    $0.75\pm0.00$       &   $0.81\pm0.00$     &   $0.77\pm0.00$       \\
K3 &    $0.67\pm0.00$  &     $0.77\pm0.01$      &  $0.85\pm0.01$      &         $0.81\pm 0.00$
\end{tabular}%
}
\label{tab:line_ours}
\end{table}
For our application, model outputs serve as a guide for mappers to accept/reject predictions, rather than as a standalone annotator of the grid. Figure \ref{fig:line_results}, shows sample line predictions (left) and a small portion of the predicted grid (right) given the ground truth labels for poles and electrical lines, in the test region.

\subsection{Comparison to a global power systems dataset}
\begin{figure}[b]
\centering
\includegraphics[scale=0.25]{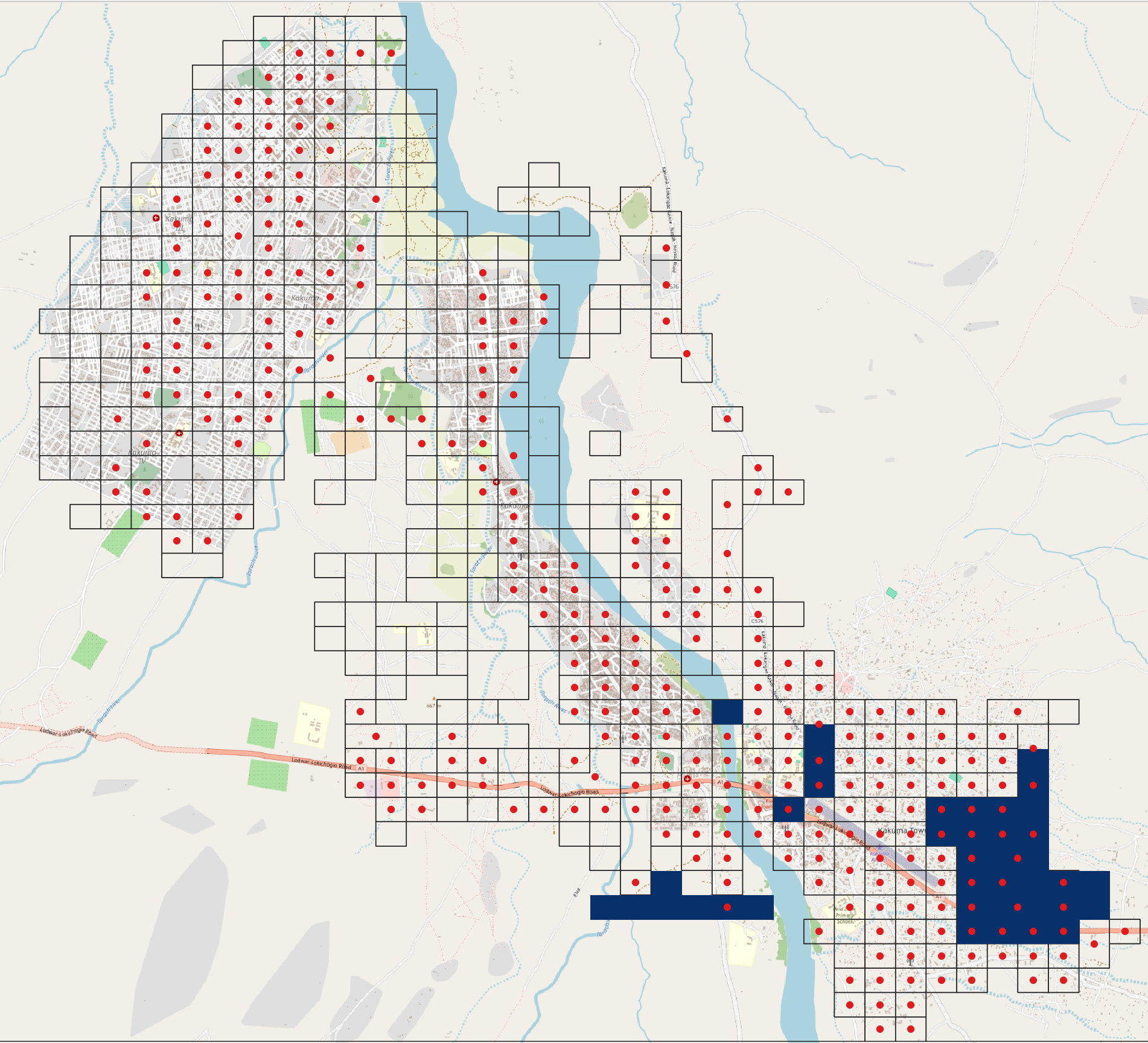}
\caption{{\sysName} provides comprehensive mapping of power distribution grids, represented by red points, covering areas that are sparsely mapped in existing datasets, shown in blue. The existing dataset, based on \cite{arderne2020predictive}, provides limited coverage of low-voltage power distribution grid for the Turkana Integrated Settlement. Grid coverage is depicted at the resolution of 250m/grid cell, demonstrating the increased granularity of our approach.}
\label{fig:newly_mapped}
\end{figure}

Information on power grid layouts is valuable for measuring electricity access and for allocating resources for grid expansion. Open access power system datasets such as the ground truth \textit{Kenya Electricity Network} dataset \cite{kenya_electricity_network_2024} provided by Kenya Power and Light Corporation (KPLC) or approximations of Low Voltage (LV) coverage such as \cite{arderne2020predictive}, provide relevant information about distribution-level grid layouts and electricity access. However, these datasets are often incomplete especially in remote areas. Our approach for reconstructing grid layouts from overhead imagery, provides a pathway for augmenting existing power grid datasets, thereby providing more comprehensive estimates of power systems coverage.  Figure \ref{fig:newly_mapped} shows the newly mapped grid coverage after running inference with our model relative to the low voltage estimate in the global power systems dataset by \cite{arderne2020predictive}.  The figure shows the presence of a power grid (red dots) using {\sysName} estimates, relative to the global power systems dataset (blue squares) for Kakuma camp at a 250m/grid cell resolution, the native resolution of the global power systems dataset. Only the Kakuma Camp is shown, as no LV data is available from open source datasets for other regions in the Turkana Integrated Settlement. With {\sysName}, we are able to include previously unmapped regions thereby augmenting existing estimates of grid topology. 
With growth in drone mapping and open imagery platforms such as OpenAerialImagery, {\sysName} can be applied across geographies to map grid topology in previously unmapped regions. 
\section{Conclusion}
\label{sec:conclusion}
In this work, we show a real-world application for mapping power grid infrastructure in unplanned settlements using high-resolution aerial imagery. By implementing advanced pole detection and line segmentation models, we show the effectiveness of {\sysName} in accurately reconstructing power grid layouts in the Turkana region, particularly within the Kakuma and Kalobeyei Camps. Results showed that {\sysName} provides a detailed and accurate representation of electricity access in these challenging environments.
The successful application of this method in resource-constrained settings and the provision of open source code, highlights its potential for broader use in other unplanned or underserved regions. Leveraging open data supports scalable solutions, providing high-precision guidance to mappers. This can greatly benefit humanitarian efforts and infrastructure planning, leading to better resource allocation, enhanced service delivery, and ultimately, improved quality of life for communities in these areas.
{\small
\bibliographystyle{ieee_fullname}
\bibliography{egbib}
}
\newpage
\section{Supplementary Material}
\subsection{Electrical pole detection illustration}
\begin{figure}[h]
\centering
\includegraphics[scale=0.45]{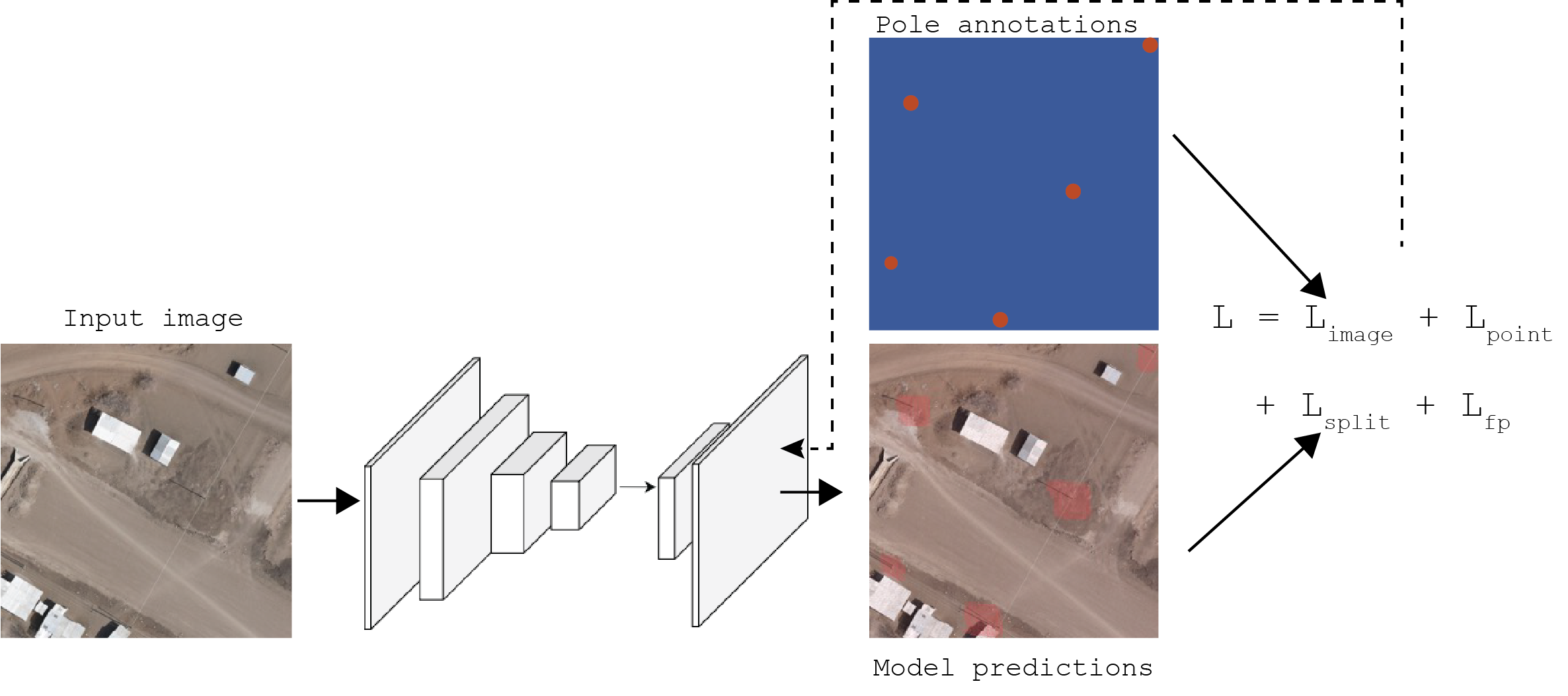}
\caption{Illustration of the pole detection workflow, including the 4 types of losses: Image level loss is an image classification loss ($L_{image}$) measuring the accuracy in classifying animage as containing a pole or not, a point level loss ($L_{point}$) for localization of poles within the image, a split level loss ($L_{split}$) to ensure unique poles are obtained and a false positive loss ($L_{fp}$) to minimize false positive detections.}
\label{fig:poledetarch}
\end{figure}

We deploy a Fully Convolutional Network (FCN8) semantic segmentation model to detect poles within the image. Figure \ref{fig:poledetarch} illustrated the point supervision method used to train the pole detection model. A mask with points is compared to the predicted blobs using the 4 component loss, to obtain a supervisory signal for training. The pole detection model is trained with the Adam optimizer at a small learning rate of 1e-6. Standard data augmentation techniques (rotations, vertical and horizontal flips, color jitter) are applied to regularize learning.

\subsection{Electrical line segmentation illustration}
\begin{figure}[t]
\centering
\includegraphics[scale=0.55]{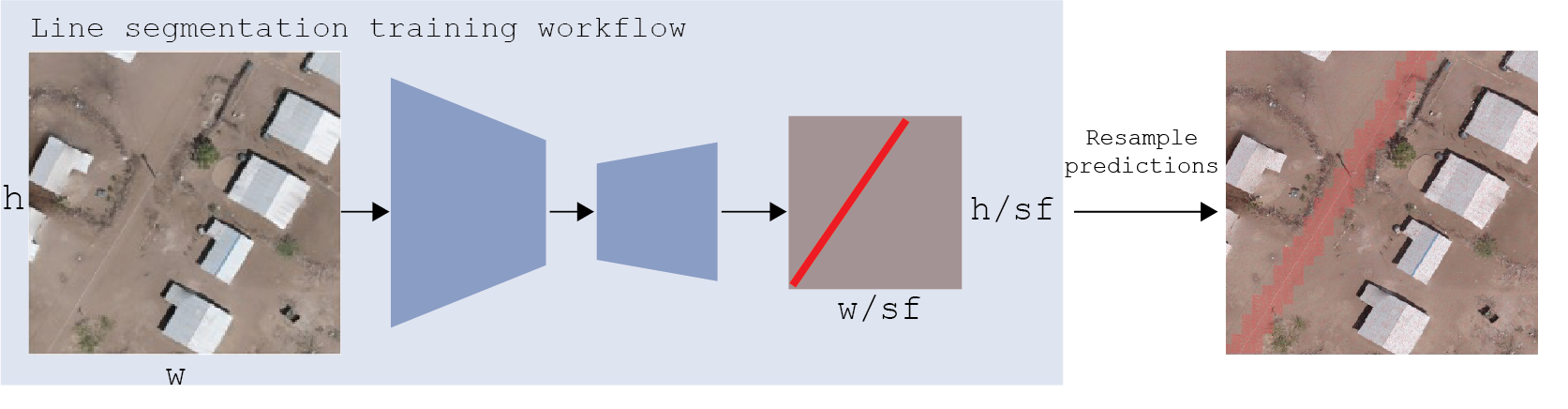}
\caption{Illustration of the asymmetric line segmentation encoder-decoder network that outputs patch-wise predictions. The downsampling factor (\textit{sf}) determines the patch sizes for predictions. Output masks are then resampled  using bi-linear interpolation to obtain a prediction mask the size of the input image.}
\label{fig:linesegarch}
\end{figure}

For the line segmentation task, we deploy an asymmetric DeepLabV3 model for patch-wise segmentation. We apply a scaling factor on the ground truth mask, to create a patch-wise mask of line presence or absence. After obtaining patch-wise predictions, the predictions are resampled using bi-linear interpolation to obtain a prediction mask the size of the input image. Figure \ref{fig:linesegarch} illustrates the line segmentation workflow used for training. The line segmentation model is trained with the Adam optimizer at a small learning rate of 1e-5. Standard data augmentation techniques (rotations, vertical and horizontal flips, color jitter) are also applied to regularize learning.
\subsection{Sensitivity to scaling factor.}
To perform patch-wise line segmentation, we experiment with three scaling factors: 1, 4 and 8. 
\begin{table}[b]
\caption{Line segmentation model performance in Kalobeyei Camp as a function of the scaling factor (\textit{sf}). The \textit{sf} of 4 produces the best localization and detection results as measured by the mIOU and F1-scores. At this \textit{sf}, the original label mask is reduced to 1/4 of its size, and the model classifies every 4x4 patch within the image as containing a line or not.}
\begin{tabular}{l|lll|lll}
\hline
           & \multicolumn{3}{c|}{\textbf{mIOU}} & \multicolumn{3}{c}{\textbf{F1-score}} \\ \hline
\textit{sf} & K1      & K2     & K3     & K1       & K2      & K3      \\ \hline
1      & 0.68    & 0.61   & 0.68   & 0.82     & 0.76    & 0.81    \\
4       &\textbf{0.70}    & \textbf{0.63}   & 0.67   & 0.82     & \textbf{0.77}    & 0.81    \\
8       & 0.69    & 0.62   & 0.68   & 0.82     & 0.77    & 0.81   
\end{tabular}
\label{tab:scalingfactor}
\end{table}
\textit{sf} of 1 leaves the ground truth label mask size unchanged, a \textit{sf} of 4 reduces the ground truth label raster to a quarter of its original size, while a \textit{sf} of 8 reduces the ground truth label raster to an eighth of its original size. 
We observe that, an \textit{sf} of 4 yields the best localization as measured by mIOU and detection as measured by the F1-score as shown in Table \ref{tab:scalingfactor}.

\subsection{Selecting a distance threshold (\textit{th}).}
\begin{figure}[t]
\centering
\includegraphics[scale=0.5 ]{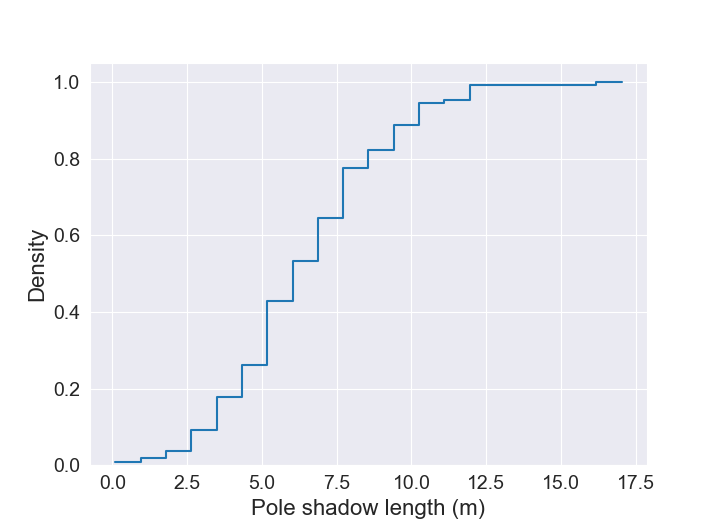}
\caption{Distribution of pole shadow lengths, given a random samples of 100 poles from the test area.}
\label{fig:pole_shadow_length}
\end{figure}

To understand the impact of the distance threshold (\textit{th}) when evaluating the pole detection model in the Turkana Intergrated Settlement, we randomly sample 100 poles from the test set and measure the shadow lengths cast by the poles. Figure \ref{fig:pole_shadow_length} shows the distribution of pole shadow lengths for the 100 sampled poles. We observe that less than 30 \% of poles cast a shadow less than 5 meters while 90\% cast a shadow of 10 meters or less. 
This observation is important because the model outputs blobs as predictions for pole locations. The centroids of the blob are then used as predicted pole locations. For poles with longer shadows and by consequence larger blobs, their centroids occur further away from the ground truth pole point location. Thus a correctly detected pole, might be classified as a false positive if the chosen distance threshold is too small. Understanding the distribution of pole shadows is also of importance so that unreasonably large distance thresholds are not selected. At very large thresholds, almost all poles would be reported as detected but these detections could be faulty as the predicted poles can occur very far from the ground truth point. By measuring the distribution of lengths cast by pole shadows in the dataset, a reasonable threshold can be selected to evaluate how well the model is performing and drive further model improvements. It is worth noting that electrical poles do not typically occur in dense clusters as they are almost evenly spaced out to support lines over a geographic area. This constraint in the physical placement of electrical poles reduces the likelihood of mismatches between predictions to ground truth pole, especially at a distance threshold less than 10m.
\end{document}